\documentclass[10pt,twocolumn,letterpaper]{article}

\usepackage{cvpr}
\usepackage{times}
\usepackage{epsfig}
\usepackage{graphicx}
\usepackage{amsmath}
\usepackage{amssymb}

\usepackage{tabularx}
\usepackage{booktabs}
\usepackage[alsoload=synchem]{siunitx}
\usepackage{flushend}

\usepackage[pagebackref=true,breaklinks=true,colorlinks,bookmarks=false]{hyperref}

\cvprfinalcopy % *** Uncomment this line for the final submission

\def\cvprPaperID{3} % *** Enter the CVPR Paper ID here

\ifcvprfinal\fi

\def\slapnicar{Slapni\v{c}ar }

\begin{document}

% %%%%%%%% TITLE
\title{Assessment of deep learning based blood pressure prediction from PPG and rPPG signals}

\author{Fabian Schrumpf\\
Leipzig University of Applied Sciences\\
{\tt\small fabian.schrumpf@htwk-leipzig.de}
% For a paper whose authors are all at the same
% institution, omit the following lines up until the closing ``}''.
% Additional authors and addresses can be added with ``\and'', just like the
% second author.
\and Patrick Frenzel\\
Leipzig University of Applied Sciences\\
{\tt\small patrick.frenzel@htwk-leipzig.de}
\and Christoph Aust\\
University Hospital Leipzig\\
{\tt\small christoph.aust@googlemail.com}
\and Georg Osterhoff\\
University Hospital Leipzig\\
{\tt\small georg.osterhoff@medizin.uni-leipzig.de}
\and Mirco Fuchs\\
Leipzig University of Applied Sciences\\
{\tt\small mirco.fuchs@htwk-leipzig.de}
\thanks{\textcopyright}}

\maketitle

% \thispagestyle{empty}

% %%%%%%%% ABSTRACT
%Their performance is often reported with respect to some mean average error (MAE) on the dataset.
%This is problematic for various reasons.
\begin{abstract}
Exploiting photoplethysmography signals (PPG) for non-invasive blood pressure (BP) measurement is interesting for various reasons.
First, PPG can easily be measured using fingerclip sensors. Second, camera-based approaches allow to derive remote PPG (rPPG) signals similar to PPG and therefore provide the
opportunity for non-invasive measurements of BP. Various methods relying on machine learning techniques have recently been published.
Performances are often reported as the mean average error (MAE) on the data which is problematic. This work aims to analyze the PPG- and rPPG-based BP prediction error with respect to the underlying data distribution.
First, we train established neural network (NN) architectures and derive an appropriate parameterization of input segments drawn from continuous PPG
signals. Second, we apply this parameterization to a larger PPG dataset and train NNs to predict BP.
The resulting prediction errors increase towards less frequent BP values. Third, we use transfer learning to train the NNs for rPPG based BP
prediction. The resulting performances are similar to the PPG-only case. Finally, we apply a personalization technique and
retrain our NNs with subject-specific data. This slightly reduces the prediction errors.
\end{abstract}

% %%%%%%%% BODY TEXT
\section{Introduction}
Blood pressure (BP) is regarded as an essential biomarker for various diseases. Techniques for discontinuous measurements are quite elaborated and comprise
auscultatory and oscillometric cuff-based methods. While they are commonly used in both clinical and home environments, they are rather unsuited for long-term
measurements due to patient discomfort and potential skin irritations.
Techniques for continuous BP monitoring are also readily available for some use cases and comprise arterial BP measurement and cuff-less sensor solutions
\cite{ding_continuous_2016, kumar_past_2020, tamura_cuffless_2020}. The former is invasive thus limited to clinical settings, the latter requires the use of multiple
sensors, e.g. ECG electrodes and PPG sensors. They actually allow continuous monitoring but are still uncomfortable for patients. Long-term measurements
even require regular recalibration using an additional cuff. Other but usually less practicable techniques to measure BP comprise ultrasound, tactile
sensor-based approaches, and vascular unloading-based methods \cite{mukherjee_literature_2018}.

In recent years research mainly focused on BP estimation from cuff-less
multi and single sensor solutions. The former often utilize time or phase
differences between different signals (usually ECG and PPG or multiple PPG)
related to the blood volume propagation through arteries
\cite{li_real-time_2020, tanveer_cuffless_2019, su_long-term_2018,
yang_estimation_2020, socrates_improved_2021}. The latter mainly exploit
morphological properties of blood volume dynamics derived from PPG measurements on a
particular site \cite{Slapnicar2019, Xing2016, schlesinger_blood_2020,
kurylyak_neural_2013,el_hajj_cuffless_2020, hsu_generalized_2020,
wang_photoplethysmography_2020, dash_estimation_2020, haddad_continuous_2020,
pandey_design_2021, han_calibration-free_nodate}. PPG-only based methods are
particularly interesting. Not only do they target single sensor
solutions, but rather is their underlying signal generation principle very much
similar to that of camera-based techniques known as rPPG. If PPG signals could
be utilized for BP estimation, a fully contactless method like
rPPG might be feasible as well, thus providing opportunities for a lot of clinical and
non-clinical application scenarios. In fact, some studies have investigated the
estimation of BP \cite{jain_face_2016, Luo2019, tran_intelligent_2020} or its
correlates \cite{Jeong2016, sugita_contactless_2019, takahashi_non-contact_2020,
finkelstein_towards_2020} from these signals.

Recent progress is driven by machine learning (ML) techniques and recently by the advent of deep learning methods. They typically pursue feature-based approaches
or perform learning in an end-to-end manner.
Feature-based methods exploit spectral or temporal properties of the PPG signal which are fed into a learning algorithm to predict systolic and
diastolic BP. End-to-end procedures leverage the waveforms themselves and implicitly derive features to predict BP. Often the accuracies of these methods
were reported to be in line with well-established standards (BHS or AAMI) \cite{stergiou_universal_2018}. It is, however, common to only report an indicator for the mean performance of
an algorithm without sufficiently taking the underlying data distributions into account. This is problematic since the performance of a learning algorithm tends
to be biased towards the mode of this distribution. That means the error appears to be small for the majority of samples but is much larger when deviating
towards the tails of their distribution. To the best of our knowledge, only one recent work studied the distribution of the BP prediction error \cite{dorr_iphone_2020}. They reported
an insufficient accuracy of the investigated method with respect to the full BP range but only divided the BP range coarsely into three intervals. This
underpins that a more detailed analysis of the error distribution is critical for the assessment of BP prediction methods in particular for clinical applications.

The BP distribution of a data set is affected by various issues. The variability of the pulse morphology among
subjects, e.g. caused by age and cardiovascular diseases, certainly affects the association of the signal shape to a particular BP. The equipment in a hospital setting differs as well and,
more importantly, the contact pressure of a PPG sensor can
affect the pulse morphology \cite{chandrasekhar_ppg_2020, allen_age-related_2020, pribil_comparative_2020}. Therefore the ability of the learning
algorithm to generalize well may be impaired. On the individual scale of a
subject, the BP variation is often limited during the measurement, e.g. due to medication (often not reported for databases), limited physical
activity or due to short record periods. This may also affect the
data distribution, particularly when training, validation and test sets are not split carefully. Lastly, the data used to train a learning model is often not publicly available. The
question which subjects were used for training and testing when
using publicly available databases remains frequently unanswered. Hence, it is challenging to assess whether an improved prediction error results from
methodological improvements or just from data selection. While these
are issues that arise for both PPG and rPPG based methods, recent works disputed
the usefulness of rPPG signals for BP estimation altogether \cite{Kamshilin2015, Moco2018}.

This paper targets (1) an empirical evaluation of the parameterization of input signals (e.g. segment cropping from the continuous signals) suited for both PPG
and rPPG based BP prediction with established neural network (NN) architectures; (2) a detailed assessment of PPG based BP prediction performance on sufficiently
small intervals of the systolic and diastolic targets; (3) rPPG based BP prediction on a data set recorded in a clinical setting based on a pre-trained and
fine-tuned network; (4) the effect of personalization by fine-tuning networks using subject-specific data.

\section{Related work}
\subsection{Deriving BP using parameterized models}
Early works for cuffless BP estimation utilize the pulse transit (PTT) or pulse arrival time (PAT). The PTT is the time delay for the pulse to travel between
two different arterial sites and the PAT is the time delay between the electrical onset (R-peak in the ECG) and the arrival of the evoked pulse wave at a
particular site \cite{zhang_pulse_2011}. They can be used to derive the pulse wave velocity (PWV) using the Moens-Korteweg equation
\cite{wippermann_evaluation_1995, callaghan_relationship_1984}. Gesche \etal used the PWV and inferred BP values by means of a linear regression model
\cite{Gesche2012}. This method resulted in a commercially available smartwatch for sleep research (SOMNOtouch NIBP, SOMNOMEDICS GmbH). Socrates \etal
\cite{socrates_improved_2021} validated this device and found a standard error of 4.2 mmHg for systolic and diastolic BP during sleep and awake phases.

\subsection{BP prediction using PPG features}
Haddad \etal used morphological features of the PPG-waveform to classify BP
into normal and hypertonic ranges using multi-linear regression \cite{haddad_continuous_2020}. Others employed dense neural networks (DNN) to derive BP from
time-based morphological features such as pulse width, pulse amplitudes and
heart rate. They also extracted features from the first and second
derivative of the PPG waveform \cite{kurylyak_neural_2013, hsu_generalized_2020, pandey_design_2021}.
Other authors used recurrent neural networks (RNN) to derive BP from time- and
frequency-based PPG-features \cite{li_real-time_2020, el_hajj_cuffless_2020, senturk_non-invasive_2020}. In \cite{tanveer_cuffless_2019, su_long-term_2018}, the authors trained a
very deep RNN by introducing skip connections between layers to overcome the vanishing gradient problem \cite{he_deep_2016}. Yang \etal constructed time series from morphological PPG features and
PTT values. They divided the time series into high- and low-frequency
components and fed them into a DNN and RNN for BP estimation \cite{yang_estimation_2020}.

\subsection{End-to-End approaches to predict BP}
\slapnicar \etal \cite{Slapnicar2019} used a parallel architecture consisting of three residual neural networks (ResNet) to
predict BP using the PPG waveform and its first and second derivatives. They also used a subject-based calibration resulting in a significantly reduced mean average error
(MAE). Schlesinger \etal used a siamese convolutional neural network (CNN) to predict BP variations
with respect to a calibration value. They used PPG spectrograms as inputs for
their NN \cite{schlesinger_blood_2020}. Baek \etal utilized ECG and PPG to derive time
and frequency domain input segments and proposed a new BP prediction model based on multiple losses\cite{baek_end--end_2019}.

Recent studies \cite{esmaelpoor_multistage_2020, eom_end--end_2020, jeong_combined_2021} used CNNs in combination with RNNs to estimate BP. The features vectors
are derived from PPG based CNN embeddings and fed into an RNN for BP prediction. Eom \etal \cite{eom_end--end_2020} achieved an MAE of
0.06 mmHg and 5.42 mmHg for systolic and diastolic BP using this approach. Wang \etal used the PPG time-course and first and second-order
derivatives as input to a parallel structure consisting of an LSTM and a CNN.
They classified BP into five different classes and achieved an overall accuracy
of 91\% \cite{wang_photoplethysmography_2020}. A more straightforward approach was pursued by Han \etal who used a
CNN with PPG-waveforms as input and achieved an overall F1-score of 0.9 when classifying BP into different hypertension classes \cite{han_calibration-free_nodate}.
Xing \etal used a simple DNN for BP estimation. They transformed the PPG signal
into the frequency domain and used its amplitude and phase spectrum as input. The mean error based on the MIMIC-II database was 0.06 mmHg and
0.01 mmHg for systolic and diastolic BP \cite{Xing2016}.

\subsection{BP estimation from rPPG}
Camera-based estimation of BP has gained significant attention in recent years. Most works derive the pulse wave from image sequences of a subject’s face from
single or multiple regions. Some methods derive rPPG just by averaging of the facial skin pixels in the green or red channel \cite{jain_face_2016, Jeong2016,
sugita_contactless_2019}. More sophisticated approaches account for light reflections and movements to reduce artefacts \cite{Luo2019, tran_intelligent_2020}.
These rPPG signals are subsequently used to estimate BP using supervised machine learning methods.

Luo \etal used transdermal imaging to derive the pulse wave from 17 different regions of interest (ROI) in the face \cite{Luo2019}. Amplitude and time domain
features were fed into a multilayer perceptron. They achieved a mean error of 0.39 mmHg and -0.2 mmHg for systolic and diastolic BP. Jain \etal derived the
pulse wave using principal component analysis (PCA) on the red color channel. Time and frequency features of the pulse signal are used to predict BP with linear
regression model. Their model achieved an MAE of 3.9 mmHg and 3.7 mmHg for systolic and diastolic BP \cite{jain_face_2016}. Other approaches translate the pulse
wave velocity and image-based pulse transit time (iPTT) concept to remote measurements. In \cite{Jeong2016} and \cite{finkelstein_towards_2020}, iPTT values
measured between the palm and the face region revealed a high correlation to PTT values based on ECG and PPG. The authors also found a moderate correlation
between the iPPT values and the BP values taken by a cuff device \cite{Jeong2016}. Many authors studied the iPTT between different ROIs in the face but only
found low correlations to BP \cite{sugita_contactless_2019, takahashi_non-contact_2020}. In \cite{sugita_contactless_2019}, largest correlations were found for
the subject’s right palm, corroborating the findings of \cite{Jeong2016, finkelstein_towards_2020}. Tran \etal used the iPTT between the face and the palm to
predict BP using a pre-trained NN. Their MAE for systolic and diastolic BP was 3.1 mmHg and 2.6 mmHg \cite{tran_intelligent_2020}.

\section{Methods}
\subsection{Datasets}
\subsubsection{PPG data}
The MIMIC-III database consists of thousands of records from various hospitals collected between 2001 and 2008 and sampled at \SI{125}{\Hz}
\cite{johnson_mimic-iii_2016}. We used a subset of the MIMIC-III database available on
Kaggle\footnote{https://www.kaggle.com/mkachuee/BloodPressureDataset}. It consists of 12000 records of PPG, ECG and ABP signals. Their authors applied extensive
preprocessing on the waveforms to provide a clean and valid dataset \cite{kachuee_cuff-less_2015, kachuee_cuffless_2017}. It represents only a
small fraction of the MIMIC-III database. but is therefore compact and contains signals with acceptable quality. However, the subject affiliation is unknown thus
rendering it unsuited for evaluating the model performance. Hence, we only used this dataset in the first part of our work to evaluate the parameterization of
input signals. It is denoted as MIMIC-A for the remainder of this paper.

A much larger portion of the MIMIC-III database was downloaded using scripts
provided by \cite{Slapnicar2019}, resulting in a total of 4000 records (PPG and ABP signal pairs). This dataset
is denoted as MIMIC-B and is used for performance evaluations.

\subsubsection{rPPG data}
Data for camera-based BP prediction was recorded in a
study at the Leipzig University Hospital. The study
design was approved by the ethics committee of the University of Leipzig.
Subjects were informed about the content of the study and gave written consent
prior to taking part. 50 subjects with a planned surgical intervention were
enrolled. After surgery, the patients were transferred to the
intensive care unit. Our recording system consisted of an industrial USB camera
(IDS UI-3040CP, 32 fps) connected to a PC (Intel NUC NUC7I7BNH). Videos of the
subjects face and upper body with approximately two hours duration were
recorded. Ground truth BP was derived from the bedside monitor with one minute temporal resolution.

\subsection{Neural network architectures}
We used three different neural network architectures to predict BP values. The first is AlexNet which is a CNN architecture originally
developed for image classification \cite{krizhevsky_imagenet_2017}. We
adopted its structure in order to use PPG time series as inputs and return
systolic and diastolic BP values instead of class predictions.

Second, we used a ResNet, i.e. a very deep CNN architecture \cite{he_deep_2016}. Their skip connections efficiently account for the vanishing gradient problem occuring in deep architectures. The ResNet was modified in the same manner as the AlexNet. The
dimensions for the input layer of these networks were \( N_{samp} \times 1 \) in the
univariate case and \( N_{samp} \times 3 \) when using the raw time series and its first and
second order derivatives. The final classification in each original model was
replaced by a regression layer consisting of two neurons for SBP and DBP with a
linear activation function.

Third, we used the architecture published by \slapnicar \etal \cite{Slapnicar2019}. It consists of a
spectrotemporal network with residual connections. It is a parallel
structure and processes PPG signals and their first and second-order derivatives.

\subsection{PPG signal processing}
\label{sec:ppg_sig_proc}
We first aimed at studying the effect of the signal length on the prediction performance. We divided the PPG and ABP signals from the MIMIC-A dataset
into segments of different lengths, i.e. 1, 2, 5, 7, 9, 11, 13, 15, 17 and 20s. This dataset is called \textit{const\_time\_xx}. Since this approach leads to an
interruption of single PPG cycles at the beginning and the end of each time window, we created an alternative dataset to evaluate whether this affects the
prediction performance. Therefore, we ensured that only full cycles (i.e. complete beats) are contained in a time window under test using the following procedure.
We estimated the heart rate based on the PPG by detecting the spectral component with the highest amplitude. Next, we divided the signal
into time segments containing an integer number of 1, 2, 5, 7, 9, 11, 13, 15, 17 and 20 PPG-waves. All segments were resampled to have equal length. It is
obvious that this also eliminates absolute temporal information. The new sampling frequency was chosen in a way that the PPG-window contains a heart rate of 60
bpm. The resulting dataset is called \textit{const\_beats\_xx}.
We processed ABP signals in a similar manner to yield datasets consisting of PPG-ABP pairs.

Recent studies showed that, apart from the raw PPG itself, derivatives yield useful information on the cardiovascular state \cite{elgendi_use_2019} and can be
useful for BP as well \cite{Slapnicar2019, hsu_generalized_2020, wang_photoplethysmography_2020}. Hence, we employed the first and second order derivatives of
each PPG-window as well and studied whether this multivariate approach reduces the BP prediction error.

We derived the ground truth systolic and diastolic BP from the ABP segments. We used a peak detection algorithm to detect systolic and diastolic peaks
\cite{elgendi_detection_2014, elgendi_systolic_2013} and derived the reference BP as the median of all peaks within each segment. We employed several plausibility checks.
All BP values outside a physiologically plausible range of 75 to 165 mmHg and 40 to 80 mmHg for systolic and diastolic BP, respectively, were discarded. Median
heart rates in each window that exceeded the ranges of 50 to 140 bpm were also rejected.

\subsection{Evaluation of NN input sequences}
\label{subsec:input_sequences}
\begin{figure*}[t]
\centering
\includegraphics[width=0.9\textwidth]{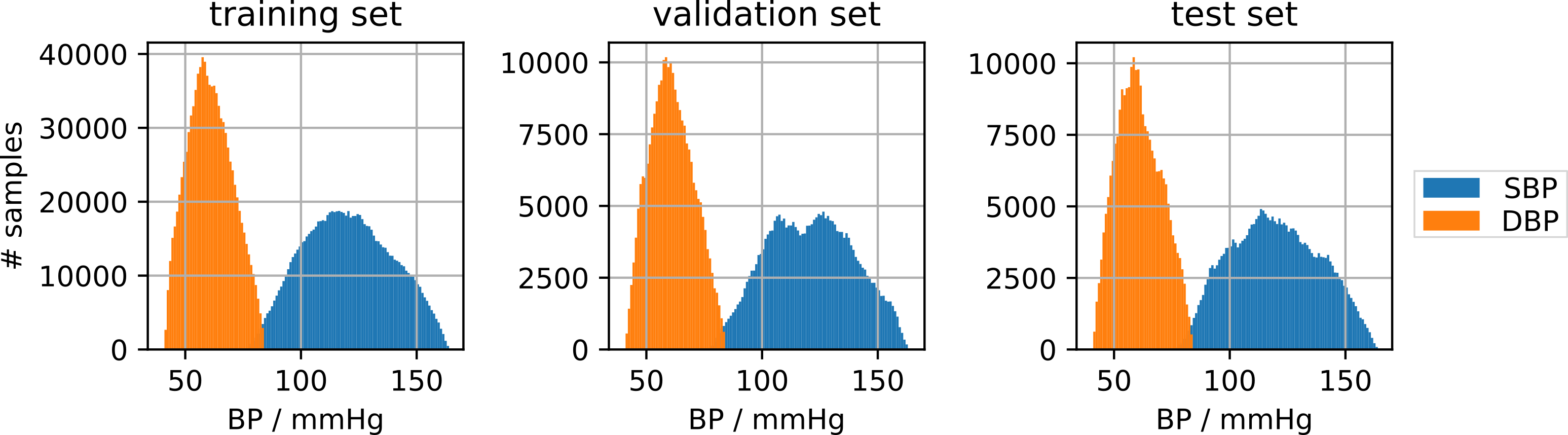}
\caption{Blood pressure distribution in the training, validation and test set in dataset MIMIC-III-B based on the MIMIC-III database}
\label{fig:BP_distribution}
\end{figure*}
We first trained our NNs with the MIMIC-A dataset to determine the proper cropping strategy and window length. We used 100000 randomly drawn samples for training the model and additional 25000 samples for validating and testing, respectively. The neural architectures and the training pipelines were implemented using
Google Tensorflow 2.4 and Python 3.5 (Adam optimizer, $\alpha=0.001$, 50 epochs, mean squared error loss). The models with the lowest validation loss were used
for subsequent performance tests.

We conducted three repeated training procedures for both AlexNet and ResNet for
each length of the input time segment under test. We then employed a paired
t-test to evaluate whether interrupting PPG cycles when cropping segments from
the continuous PPG signal affects the MAE. We conducted this analysis separately
for the AlexNet and ResNet architecture.

To determine the optimal window length we employed a twofold strategy. First, we evaluated the NN performances with respect to the input length of the segments
derived from PPG signals. Second, we evaluated how a particular length would affect the SNR of rPPG segments. We aimed at selecting a length that, on the one
hand, ensures high SNR values (thus resulting in longer segments) but, on the other hand, maximizes the number of segments (and therefore enforcing shorter
segments) available for NN training. The latter is motivated by the fact that various studies have shown that a large number of training examples is crucial for
the successful training of NNs. Hence, our procedure aimed at finding a trade-off resulting in a sufficiently large number of training examples with an acceptable SNR for rPPG, given
the limited amount of samples in the rPPG dataset. As before, we divided the rPPG dataset into windows of different length and calculated the SNR for every time
window. We used a threshold of \SI{7}{\dB} SNR above which a rPPG segment would be accepted. We finally used a the shortest window length that provides a large
number of acceptable time windows while still being long enough to expect good results for BP prediction with the NN.

\subsection{PPG-based BP prediction}
\label{sec:PPG_Training}
\subsubsection{Data preprocessing}
\label{subsec:data_processing}
The MIMIC-B dataset was divided into windows using the optimal cropping strategy and window length determined in section \ref{sec:ppg_sig_proc}. Since this data came directly from
the physionet.org database, additional preprocessing steps had to be applied. First, a 4th order Butterworth band-pass filter was applied to the PPG-signals.
Cut-off frequencies were set to \SI{0.5}{\Hz} and \SI{8}{\Hz}. Second, the signal-to-noise ratio (SNR) was calculated for all signals \cite{DeHaan2014a}. All signal windows with an SNR
below \SI{-7}{\dB} were discarded. All PPG-windows were normalized to zero mean and unit variance.

\subsubsection{NN training and evaluation}
The dataset was split into training, validation and test set on a subject-basis to prevent contamination of the validation and test set by training data. We
used 3750 subjects for training and 625 subjects for validation and testing. Among these subjects, we have randomly drawn 1.000.000 samples for
training, 250.000 samples for validation and 250.000 samples for testing. The BP distribution among these datasets can be seen in Fig. \ref{fig:BP_distribution}.

Input pipelines and NNs (AlexNet, ResNet, model from \slapnicar \etal) were implemented using Google TensorFlow 2.4 and Python 3.8 was used for training (Adam
optimizer, $\alpha=0.001$, euclidian loss, 60 epochs). We used the models with the lowest MAE on the validation set for further testing.

For evaluation purposes, we additionally used a mean regressor that always predicts the systolic and diastolic BP from the
training set. A well generalizing ML method will exceed the mean regressor's performance.
We used the MAE metric to asses the performance of all methods. In contrast to other work, we determined the prediction errors both for the full dataset and
separately for rather narrow BP bins, altogether spanning the range of ground truth BP values contained in the datasets.

\begin{figure}[b!]
\centering
\includegraphics[width=\columnwidth]{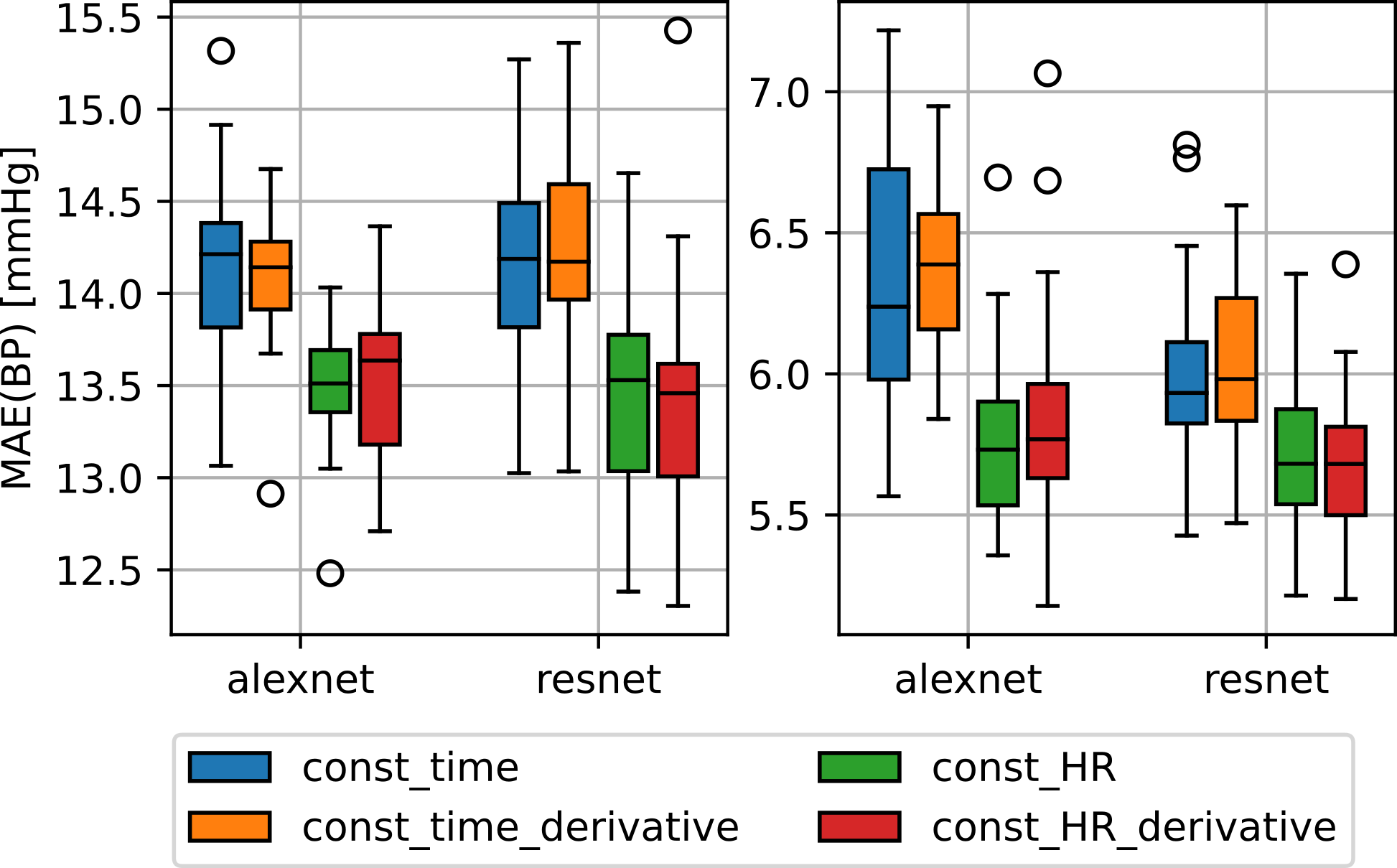}
\caption{Difference in MAE between \textit{const\_time} and \textit{const\_HR} datasets for AlexNet and ResNet. (a): MAE for the systolic blood pressure; (b): MAE for diastolic blood pressure; \textit{const\_time\_derivative} and  \textit{const\_HR\_derivative} datasets include the first and second order derivatives of the PPG windows contained in the datasets \textit{const\_time} and \textit{const\_HR}.}
\label{fig:mimic_a_compare_datasets}
\end{figure}

\subsection{rPPG-based BP prediction}
\subsubsection{Preprocessing}
ROIs on a subject’s forhead and cheeks were labelled manually after the recording using custom software. We used the Plane-orthogonal-to-skin (POS) algorithm to derive the pulse wave from
the skin pixels \cite{Wang2017}.  The resulting rPPG-signal was then inspected visually. Twenty-five subjects with heavy motion artifacts, frequent movement or
insufficient lighting were deemed unsuitable and excluded from further analysis. Remaining data was divided into windows based on their heart rate. Seven
heartbeats were included (compare Sec. \ref{subsec:input_sequences} and \ref{subsec:res_input_sequence}) and each window was resampled according to the
procedure for the MIMIC-B dataset (Sec. \ref{subsec:data_processing}). We calculated the SNR value for each rPPG window according to \cite{DeHaan2014a} and excluded
windows with an SNR below \SI{-7}{\dB}. Ground truth BP values	 were downloaded from the bedside monitor.

\subsubsection{Transfer learning}
The resulting dataset was used to fine-tune the pre-trained NNs. Transfer learning exploits the idea that rPPG and PPG waveforms share similar properties and
should therefore give rise to similar relevant features during NN training. Due to the low amount of data, however, an entire retraining of these NNs
was not feasible. We therefore only optimized the final layer while freezing all other network weights. We used the Adam optimizer ($\alpha=0.001$) and
fine-tuned until the MAE stopped decreasing. The best model given by leave-one-out cross-validation was then used to evaluate the model using the test subject.
We also investigated whether the personalization strategy presented in \cite{Slapnicar2019} improves the prediction accuracy. In addition to the training
dataset we reserved 20\% of the test subject’s data for training and validated the model with the remaining 80\%.

\section{Results}
\subsection{PPG-based BP prediction}
\subsubsection{Input signals}
\label{subsec:res_input_sequence}
Figure \ref{fig:mimic_a_compare_datasets} shows the training results of the AlexNet and ResNet architectures with the \textit{const\_HR} and
\textit{const\_time} datasets with and without additional time derivatives. Note that we combined the results obtained for all variations of the segment lenghts
for this analysis. The prediction errors based on \textit{const\_HR} are lower in comparison to \textit{const\_time}. The significance of these findings was
confirmed using a paired t-Test ($p < 0.01$). The use of the first and second order derivatives shown as \textit{const\_HR\_derivative} and
\textit{const\_time\_derivative} did not yield to a general improvement compared to the univariate case. Just the SBP MAE of the AlexNet was slightly lower than
in the univariate case. Due to this and for the sake of simplicity, we did not consider derivatives for our further analysis.

\begin{figure}[t!]
\centering
\includegraphics[width=0.7\columnwidth]{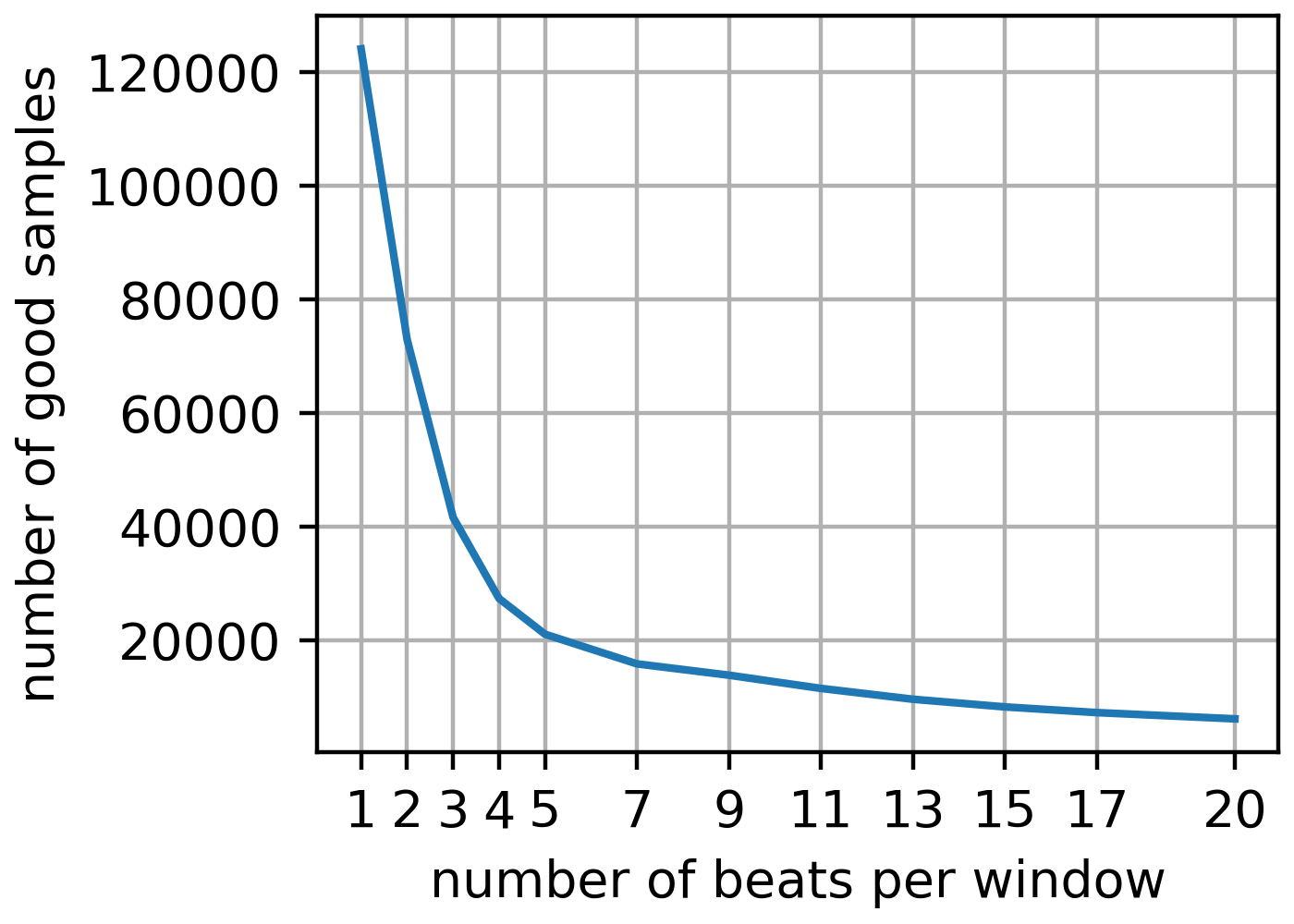}
\caption{Fraction of acceptable samples in the rPPG dataset with respect to the window length. Samples with an SNR below the threshold of -7 dB were discarded.}
\label{fig:mimic_a_good_samples}
\end{figure}
According to Sec. \ref{subsec:input_sequences}, we aimed to derive a suitable length of the input signal as a trade-off from PPG and rPPG data. First, we
evaluated the prediction error with respect PPG input signals. We expected a slight increase of the error towards longer segments since the morphology of PPG
cycles slightly varies over time and therefore would introduce undesired ambiguities with respect to the underlying BP. However, our empirical analysis (3
repetitions for each NN and length parameter) did not confirm this effect and resulted in an almost equal prediction error for each tested length. Note that
because of the high computational effort that would be necessary to obtain a sufficiently high number of repeated training procedures for each NN and length
parameter (30+ each), a statistical justification of this effect can not be provided. From this perspective, the longest possible segment length (\SI{20}{\s}) seemed
useful.

Second, we analyzed the SNR of rPPG segments with respect to their lengths (Fig. \ref{fig:mimic_a_good_samples}) and aimed to maximize the number of resulting
samples available for training. It should be noted that the SNR measure becomes less appropriate towards small segment lengths which also explains the sharp
decline at the beginning of the curve. It seemed reasonable to select a value beyond the bend of the curve to account for this effect. We finally selected \SI{7}{s} as
the segment length for all further analyses.
\begin{figure}[t!]
\centering
\includegraphics[width=\columnwidth]{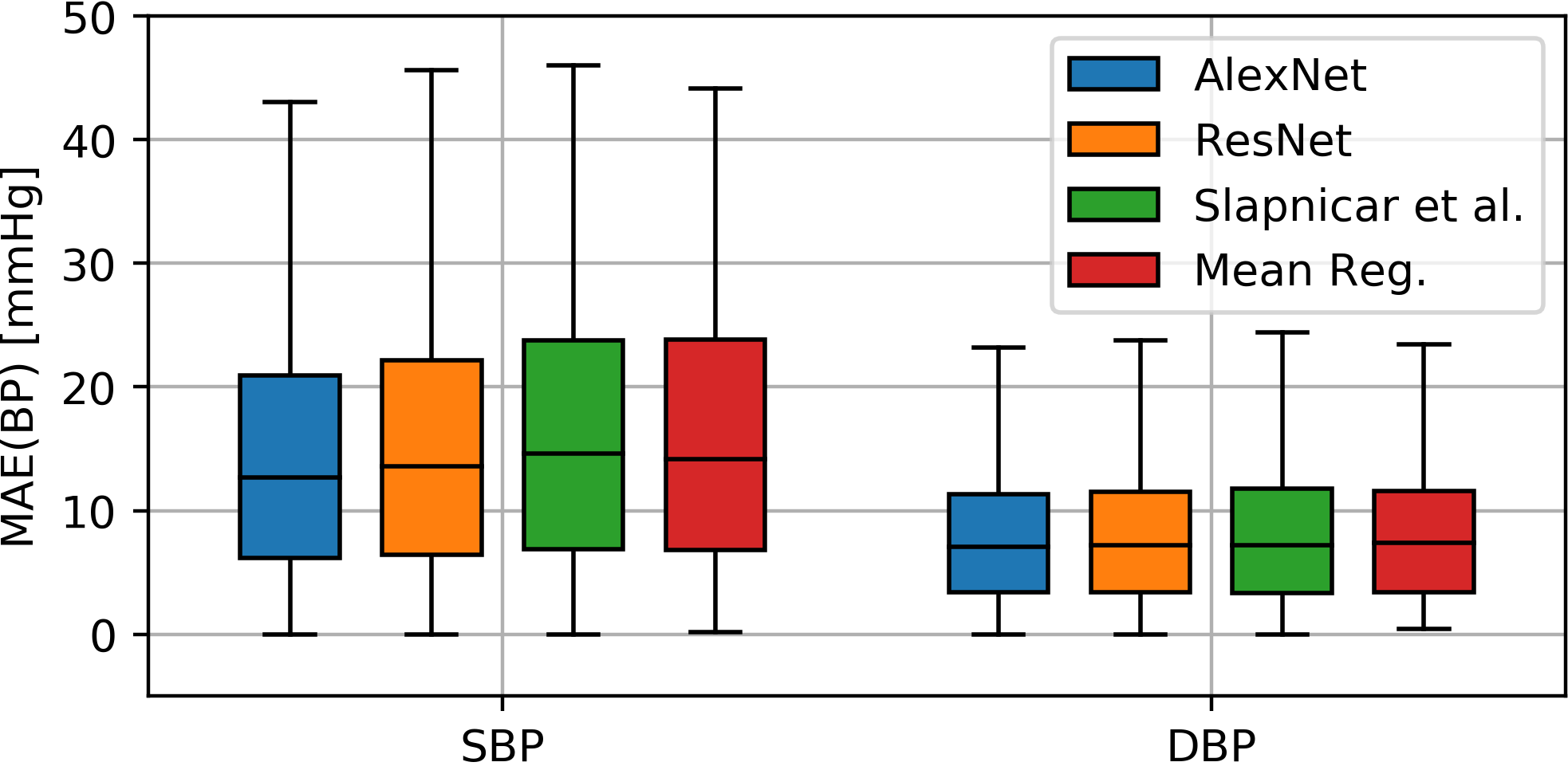}
\caption{Comparison of the MAE for SBP and DBP of the three analyzed CNN architectures in comparison to the mean regressor.}
\label{fig:mimic_a_compare_architectures}
\end{figure}

\begin{figure}[b!]
\centering
\includegraphics[width=\columnwidth]{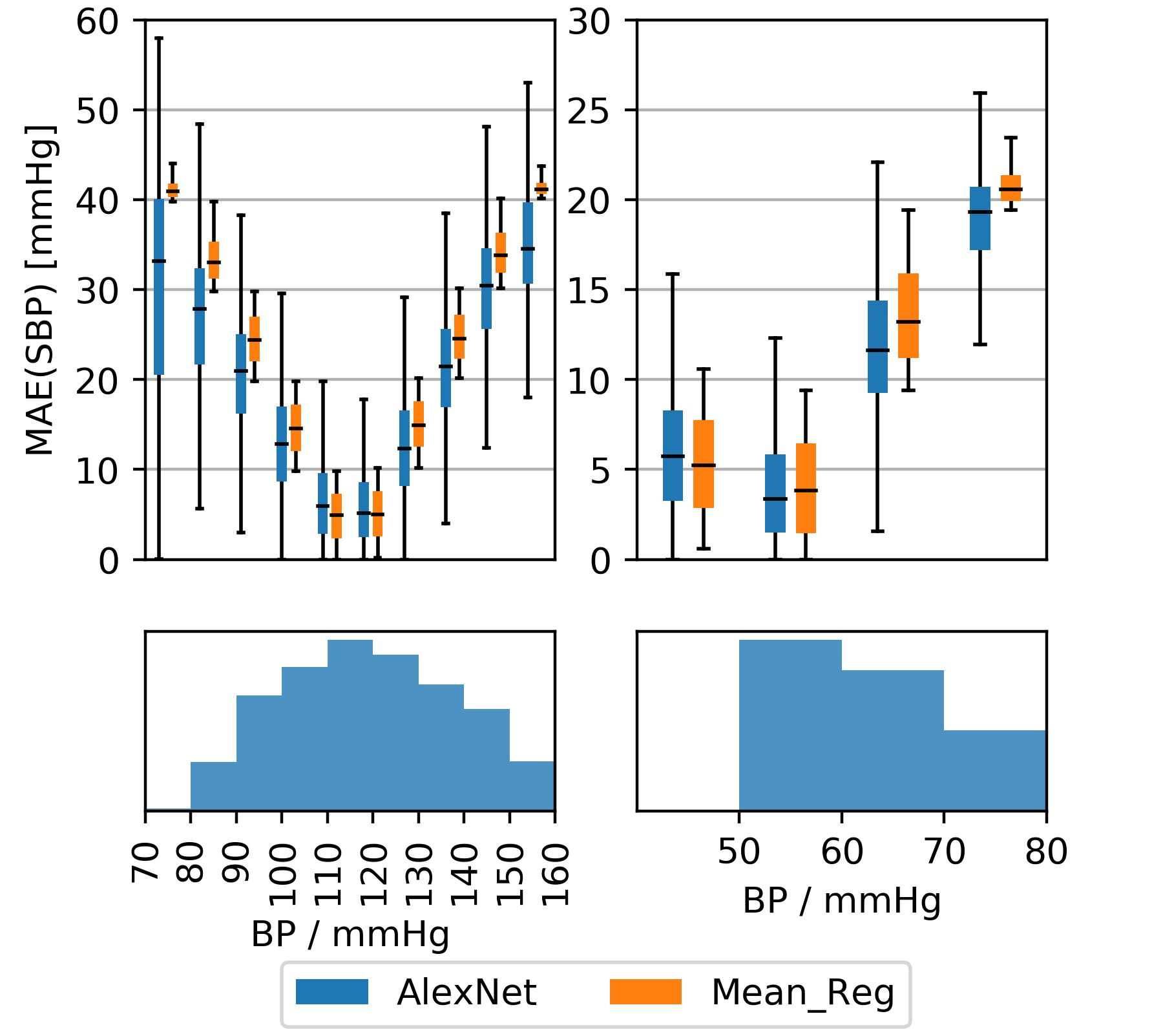}
\caption{MAE depending on the ABP-based blood pressure for the AlexNet architecture. The admissible blood pressure range was divided into bins of width 10 mmHg. For reference, the distribution of SBP and DBP is displayed at the bottom of the plot.}
\label{fig:mimic_a_binned_error}
\end{figure}

\subsubsection{Predicting BP using PPG data}
Results after training the CNN architectures with the bigger MIMIC-B dataset can be seen in Fig. \ref{fig:mimic_a_compare_architectures}. It is evident that
the DBP MAE is lower than the SBP MAE. This can be attributed to the fact that its range of values is smaller than the SBP’s range. We compared the MAE of the
AlexNet, ResNet and the model from \slapnicar \etal to the MAE of the mean regressor using a paired t-Test. Figure \ref{fig:mimic_a_compare_architectures} shows
that particularly the DBP errors of the NNs were very similar to those of the mean regressor. However, the DBP MAE of the AlexNet and ResNet architectures were
significantly lower than the mean regressor. This could not be found for the model \slapnicar \etal. When analyzing the SBP MAE, we discovered that AlexNet and
ResNet achieved significantly lower errors than the mean regressor.

To analyze the BP in dependence of the MAE we divided the acceptable BP input range into bins of \SI{10}{\mmHg} width and calculated the error for each bin
separately for each architecture. Since the results are very similar, only those for the AlexNet are shown (Fig. \ref{fig:mimic_a_binned_error}). The error
strongly varies across BP bins thus emphasizing a large dependence of the error on the underlying BP.
Each architecture achieves the lowest MAE in a range of 100 - 120 mmHg (SBP) and 56 - 60 mmHg (DBP). The BP distribution of the test set is shown at the bottom
of Fig. \ref{fig:mimic_a_binned_error}. It can be seen that the error is inversely proportional to the number of samples in the respective bin. This suggests
that the ML methods achieve a training loss reduction by concentrating their predictions on the BP range with the most frequent training examples.

\begin{figure}[t!]
\centering
\includegraphics[width=\columnwidth]{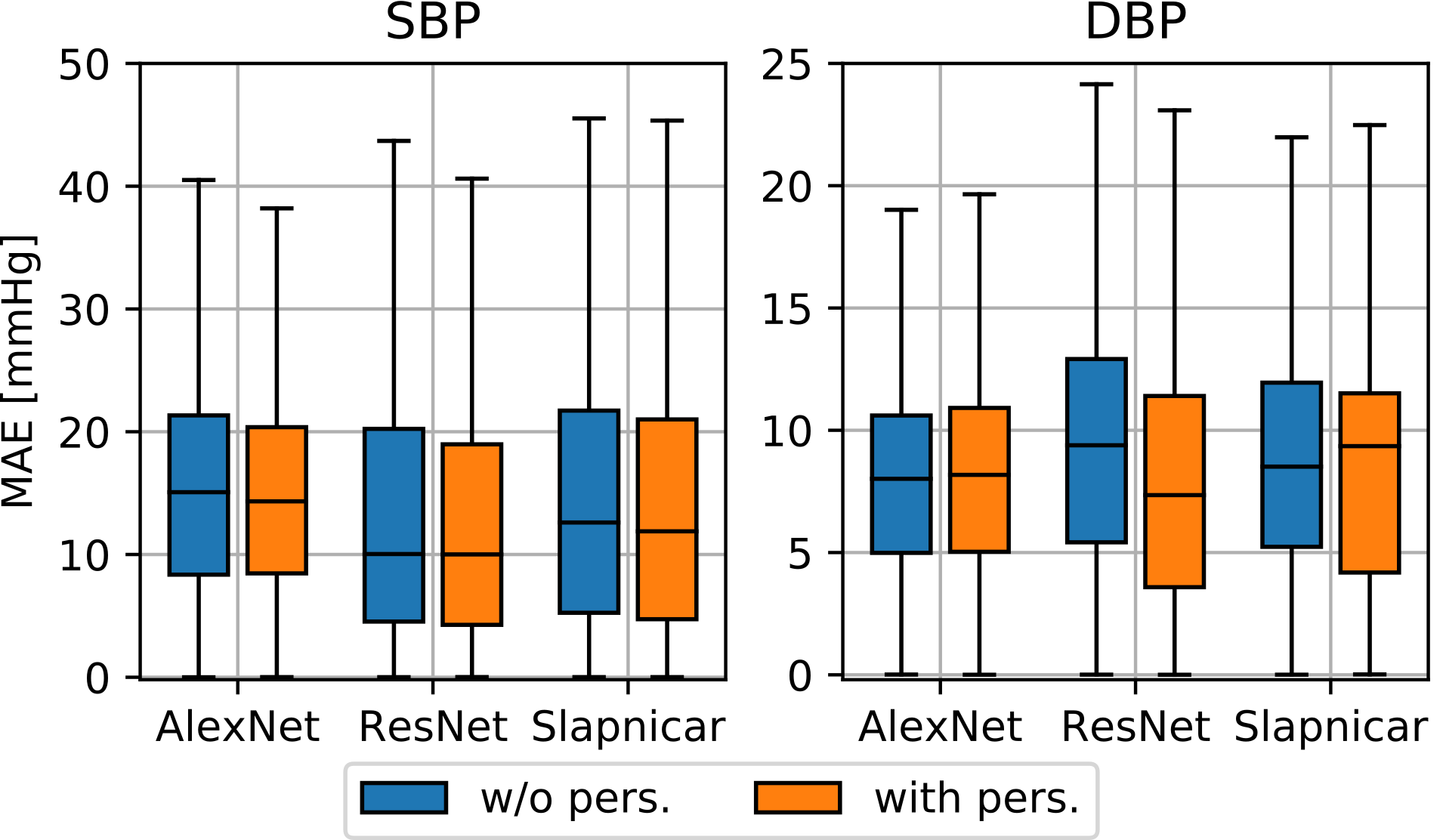}
\caption{Comparison of the MAE between fine tuning the neural networks with and without personalization.}
\label{fig:rppg_pers}
\end{figure}

\subsection{rPPG based BP prediction}
Next, we fine-tuned the NNs trained on PPG data for BP prediction using rPPG data. From 14 subjects included in the analysis, 12 were used for training, 1 for validation and 1 for testing.

Table \ref{tab:rppg_pers} (top) shows the overall MAE and STD after fine-tuning with rPPG data without personalization. It can be seen that the ResNet model
achieves the lowest MAE in terms of SBP. AlexNet achieves the lowest DBP MAE. The results of the model from \slapnicar \etal are close to the mean regressor.
Figure \ref{fig:rppg_pers} (blue boxes) shows a boxplot of the MAE based on rPPG signals. The differences in SBP MAE and DBP MAE are statistically significant (\(p
\leq 0.05\)). This emphasizes that the results may be very different on a per-subject basis but the overall performance of the three NNs is quite similar both
among each other and in comparison to the mean regressor.

Finally, we compared the impact of personalization on the BP prediction error after fine-tuning the NN. Figure \ref{fig:rppg_pers} (red boxes) and Tab. \ref{tab:rppg_pers} (bottom)
show that personalization has the biggest effect on the AlexNet and ResNet architectures. The SBP MAE improved by 0.5 mmHg and 0.51 mmHg, respectively. The
DBP MAE of the ResNet improved by 1.29 mmHg. The SBP MAE of the \slapnicar \etal network improved by 0.48 mmHg. However, changing the training/test split
also influences the performance of the mean regressor. Its SBP MAE increased by 0.5 mmHg while the DBP MAE decreased by 0.4 mmHg.
Hence, the model from \slapnicar \etal performed slightly better than the mean regressor without personalization and is on par with
the mean regressor after personalization.
\begin{table}[t!]
\centering
\begin{tabularx}{\columnwidth}{|c|X|X|X|X|}
%\toprule
\hline
&\multicolumn{2}{c|}{SBP [mmHg]}&\multicolumn{2}{c|}{DBP [mmHg]}\\
%\toprule
\hline
\hline
& Mean & Std & Mean & Std\\
%\midrule
\hline
\multicolumn{5}{|l|}{\textbf{no personalization}}\\
%\midrule
\hline
AlexNet& 15.7&9.66&8.27&4.67\\
ResNet&13.02&10.55&9.81&9.88\\
\slapnicar \etal&14.04&9.82&8.64&4.77\\
Mean Reg.&13.4&10.1&8.9&5.2\\
%\midrule
\hline
\multicolumn{5}{|l|}{\textbf{with personalization}}\\
\hline
%\midrule
AlexNet&15.2&9.11&8.52&4.92\\
ResNet&12.51&12.61&8.3&9.84\\
\slapnicar \etal&13.56&10.12&8.5&5.02\\
Mean Reg.&13.9&10.4&8.5&5.0\\
\hline
%\bottomrule
\end{tabularx}
\caption{MAE of the neural networks after fine-tuning the pre-trained networks using rPPG-data. Top: Fine-tuning \textit{without} personalization; Bottom: Fine-tuning \textit{with} personalization using 20\% of the test subject’s data for training.}
\label{tab:rppg_pers}
\end{table}

\section{Discussion}
This paper investigated the feasibility of BP estimation using PPG- and rPPG-based pulse wave signals. Importantly, our aim was not to derive a particularly
accurate model to achieve state-of-the-art performance for BP prediction from PPG but rather (1) to explore the dependence of these models on some important time domain
properties of the PPG and rPPG input signal, (2) to learn how these models perform not only in terms of a mean performance on a given dataset but on a more fine
grained scale of multiple BP bins and (3) to investigate the feasibility of transfer learning from PPG for rPPG based BP prediction. We also ensured a careful
split of the data and divided the dataset based on subjects to avoid contaminating the validation and test data with training data.

First, we conducted an empirical evaluation of the parameterization of the input signals that were suited to train our NNs. In particular, we analyzed the
window length, cropping of segments from continuous signals and the use of derivatives. We used established NN architectures (i.e., AlexNet and ResNet) from the
literature and adopted them for BP prediction. Moreover, we used the architecture as presented by \slapnicar \etal which is optimized for BP prediction from
PPG. We found that the use of derivatives does not provide significant improvements and is less important. Avoiding the introduction of strong phase
discontinuities using an appropriate cropping of segments from continuous signals significantly improves the prediction performance. Finally, the
total length of the input sequence was less important with respect to the PPG-based prediction errors. The selection of the length of segments was
mainly driven by rPPG data to maximize the number samples in the dataset.

Second, our analysis of the BP dependent prediction error reveals that the NNs are partly superior over the mean regressor. However, effect sizes were small.
Besides that, we found a strong dependence of the bin-wise prediction error on the number samples in the particular BP bin in the underlying distribution. The
most accurate predictions occurred in BP bins containing the most samples. This emphasizes the tendency towards predicting the mode of the training dataset. A
similar dependence was found in \cite{dorr_iphone_2020}, which led to the retraction of a publicly available smartphone app for BP prediction. However, their
results are based on a much coarser subdivision of the BP range.

In order to compare our results to previous work, we additionally evaluated the mean performance. None of the results met the requirements as defined in the
relevant BHS and AAMI standards, which require the probability of a BP measurement device to provide an acceptable error (\(BP \leq \SI{10}{\mmHg}\)) to
exceed 85\% \cite{stergiou_universal_2018}. Especially the high MAEs in the higher and lower BP ranges pose a problem since many clinical applications rely on
an acceptable accuracy in hypo- and hypertensive ranges. Importantly, our BP MAE is in accordance with \slapnicar \etal \cite{Slapnicar2019} who also
accounted for subject specific affiliations to training and testing datasets. Note that we did not employ any hyperparameter tuning which might further improve
our results, but certainly only on a gradual scale. In contrast, other authors who did not explicitly mention a subject-based dataset split reported
substantially lower prediction errors \cite{chandrasekhar_ppg_2020, allen_age-related_2020, pribil_comparative_2020, mceniery_normal_2005}. These differences
suggest that morphological inter-individual variations due to age, comorbidities, medication and measurement equipment prevent the investigated NNs to
generalize well. Recent works by Zhang \etal addressed this issue using neural architectures specifically tailored to learn domain-invariant features
\cite{zhang_developing_2020}. This was beyond the scope of our study since we focused on NN architectures which are already applied on a broad basis.

Our study emphasizes the following: To develop an ML technique that aims to be of relevance for practical clinical applications, one must (1) evaluate the
prediction error of a model over the full systolic and diastolic BP range and (2) therefore carefully take the data distributions in training and test sets into
account. While this is of course obvious and an important rule for designing ML algorithms in general, it is not yet treated with the necessary care among published literature for ML based BP prediction. In particular, differences in age, the health state and medications are likely to have a strong
influence on the BP distribution in a dataset. It is also worth to further investigate the relation between the PPG morphology and the sensor contact pressure
\cite{chandrasekhar_ppg_2020} which might have an even more severe effect on the ability of an ML model to generalize at all. This is especially the case when using
public databases where patient records can emerge from various sources. Given our findings, it does not seem unlikely that the mentioned and probably even more issues causing variations in the PPG morphology might render BP predictions from PPG a highly ill-posed problem for real world applications. An appropriate split into subject-dependent training and test sets must be ensured in any case. Including data of
training subjects in the test set by randomly splitting the datasets on a sample-basis leads to an overestimation of the model performance.

Third, we fine-tuned our NNs using rPPG data conducted in a clinical study. We found that the estimation error greatly varied between subjects. Given the
ResNet, the overall error was still smaller than the mean regressor. Like for PPG, we also investigated the effect of personalization. We discovered a
considerably improved prediction accuracy in comparison to the mean regressor. We emphasize that we had only limited training data for fine-tuning and therefore
could only fine-tune the final layer of each NN. Given more data, it would be conceivable to tune additional layers and possibly enhance the prediction
performance.

Remote measurement of BP using standard RGB cameras is still an active field of research. While it still seems questionable to derive BP from PPG-only data, it
remains even more questionable whether rPPG data is actually suitable for BP estimation \cite{Kamshilin2015, Moco2018, Moco2018c}. Future studies should
concentrate on end-to-end approaches since the rPPG-signal’s low SNR hampers methods based on morphological features, especially when different skin tones,
motion and changes in illumination are involved \cite{moco_skin_2016, lin_study_2017}.

\section*{Acknowledgements}
This work was funded by the German Federal Ministry of Economics and Technology (BMWi) (FKZ 49VF170043).
{\small
\bibliographystyle{IEEEtran}
\bibliography{CVPM_2021}
}

%Figure \ref{fig:rppg_pers} shows the MAE on each of the test subjects in comparison to the mean regressor. We point out that the training set
%%changes during each step of the leave-one-out cross-validation scheme, leading to varying performances of the mean regressor in terms of MAE on each
%test-subject. MAE per subject present an ambiguous picture. While the neural networks show an MAE comparable to the mean regressor for most subjects, there is
%no clear trend for any investigated algorithm. Every method works well for some subjects and worse for others.

\end{document}